\documentclass[11pt]{article}

\usepackage{acl}

\usepackage{times}
\usepackage{latexsym}

\usepackage[T1]{fontenc}

\usepackage[utf8]{inputenc}

\usepackage{microtype}

\usepackage{inconsolata}

\usepackage{graphicx}
\usepackage{booktabs}
\usepackage{amsfonts}
\usepackage{amsmath}
\usepackage{comment}
\title{Lightweight but Linguistically Fair?\\
Quantisation Robustness across Bavarian, German, and English}

\author{
  \textbf{Jophin John\textsuperscript{1}}
  \quad
  \textbf{Michael Hoffmann\textsuperscript{1}}
\\
  \textbf{Piroska Lendvai\textsuperscript{2}}
  \quad
  \textbf{Florent Dufour\textsuperscript{3}}
\\[0.7em]
  \textsuperscript{1}Leibniz Supercomputing Centre (LRZ), Garching, Germany
\\
  \textsuperscript{2}Bavarian Academy of Sciences and Humanities, Munich, Germany
\\
  \textsuperscript{3}TUM University Hospital and Technical University of Munich, Munich, Germany
\\
 {\small \textbf{Correspondence:} \href{mailto:Jophin.John@lrz.de}{Jophin.John@lrz.de}}
}

\newcommand{\bavground}{\textsc{BavGround}}
\newcommand{\genba}{\textsc{GENBA-10B}}

\begin{document}

\maketitle

\begin{abstract}
Post-training quantisation reduces the memory cost of large language models,
but it remains unclear whether compression preserves language skills of regional
varieties as reliably as their standard-language counterparts. We study this
question for Bavarian, Standard German, and English using \bavground{}, a multiple-choice benchmark, where
206 source questions are aligned across all three languages. Across 14
open-weight 7B--10B instruction models, we compare BF16 with 8-bit GPTQ and
two 4-bit settings, GPTQ and AWQ. Eight-bit GPTQ remains close to BF16 and
largely preserves the relative behaviour of the three languages, while 4-bit
quantisation was found to be much less stable. The two 4-bit methods reduce average
accuracy by about 1.4--1.5 percentage points and change 11.9--16.0\% of BF16
decisions, with changes concentrated on examples where the BF16 model is
already uncertain. This instability does not translate into a consistent
Bavarian disadvantage. Although Bavarian, a low-resource dialect, has lower baseline accuracy than
German and English, the language most affected at 4 bits varies across models
and quantisation methods. Calibration further shapes these cross-language effects: the relative
Bavarian--German effect changes with calibration language, while changing the
calibration source has little average effect on GPTQ-4 but can affect AWQ-4
much more strongly. Free-form evaluation on five models further shows
that 4-bit compression can degrade generated responses even in settings where
multiple-choice accuracy changes only modestly, while 8-bit GPTQ remains
comparatively stable. Overall, our results show that quantisation robustness
for regional language varieties depends on the model, quantisation method,
calibration data, and evaluation setting, and cannot be measured by
aggregate multilingual accuracy alone.
\end{abstract}

\section{Introduction}

Post-training quantisation is increasingly used to reduce the memory cost of
large language models and enable deployment under tighter hardware budgets.
Weight-only methods such as GPTQ \citep{frantar2023gptq} and AWQ
\citep{lin2024awq} can substantially reduce model size while often preserving
aggregate benchmark accuracy. This makes quantisation particularly attractive
for open-weight models and deployment settings in which memory capacity is a
primary constraint.

However, preservation of aggregate performance does not imply that all
languages are preserved equally well. Multilingual models are trained and
evaluated disproportionately on high-resource standard languages, while
regional and low-resource varieties often receive less representation in
training data, tokeniser construction, and benchmark design
\citep{ahia2021lowresource,hooker2021compresseddeepneuralnetworks}. Prior work
has shown that quantisation can affect languages unevenly and that automatic
metrics can underestimate degradation visible in generated outputs
\citep{marchisio2024quantizationmultilingual}. What remains less understood is
whether these effects also extend to regional varieties that are
linguistically close to a high-resource standard language.

Answering this question requires a comparison in which language varies while
the underlying content remains fixed. Standard multilingual benchmarks such as
MMLU \citep{hendrycks2021measuringmassivemultitasklanguage}, MGSM
\citep{shi2023mgsm}, and FLORES-200 \citep{costajussa2022flores} do not provide
the aligned regional-variety--standard-language setting needed for such a
comparison. Without this alignment, differences between language varieties
can be difficult to separate from differences in the questions or evaluation
conditions themselves.

We address this gap using \bavground{}~\citep{john2026bavgroundbenchmarkregionalcultural}, a multiple-choice benchmark, whose 206 source questions are aligned
across Bavarian, Standard German, and English. This allows us to compare each
quantised model with its own BF16 baseline on the same underlying questions
and examine whether compression affects Bavarian differently from German and
English. We evaluate 14 open-weight 7B--10B instruction models under BF16,
GPTQ-8, GPTQ-4, and AWQ-4, and further test how these cross-language
effects change with both the language composition and source of the
calibration data.

Across this controlled comparison, GPTQ-8 remains close to BF16 and largely
preserves the relative performance of Bavarian, German, and English. At
4 bits, robustness becomes more dependent on the model, quantisation method,
and calibration data, without consistently disadvantaging Bavarian.
Free-form evaluation further reveals degradation that is not always apparent
from multiple-choice accuracy.

Our contributions are threefold:
\begin{itemize}
\item We conduct a controlled quantisation study across 14 open-weight
7B--10B models using aligned Bavarian, Standard German, and English questions,
allowing direct comparison of a regional variety with its standard language
across precision levels and quantisation methods.
\item We examine how calibration data shape cross-language quantisation
robustness by varying both their language composition and source, and relate
item-level prediction instability to BF16 decision margins.
\item We extend prior findings on evaluation-dependent multilingual
quantisation effects to a dialect--standard-language pair. Our blinded
free-form evaluation tests whether robustness observed in multiple-choice
accuracy carries over to generated responses.

\end{itemize}

\section{Related Work}

\paragraph{Post-training quantisation:}
Post-Training Quantization (PTQ) \citep{frantar2023gptq,lin2024awq,xiao2023smoothquant} converts the
weights and, in some approaches, the activations of a neural network from
their original 32- or 16-bit floating-point representations to lower-precision
formats, typically 8-bit or 4-bit integers. GPTQ uses approximate second-order
information to minimise layer-wise reconstruction error during low-bit weight
quantisation \citep{frantar2023gptq}, whereas AWQ identifies
activation-salient channels and rescales them prior to weight quantisation
\citep{lin2024awq}. These methods can provide strong quality--efficiency
trade-offs, with prior evaluations showing that moderate quantisation often
preserves much of a model's aggregate performance, particularly for larger
models \citep{giagnorio2026evaluating}.

Early studies established that neural networks can retain useful
performance under reduced numerical precision, while highlighting
their sensitivity to quantisation \citep{sung2015resiliency}.
Aggregate performance, however, does not always change uniformly with
precision. Quantised models can sometimes match or even slightly outperform
their higher-precision counterparts on particular tasks or benchmarks
\citep{badshah2024quantifying,lee2025exploring}. Such improvements tend to
depend on the model, task, and evaluation setting rather than reflecting a
general benefit of lower precision. More importantly, stable or improved
aggregate scores can conceal losses on particular subsets of the evaluation
data. This makes it necessary to examine systematically where quantisation changes model
behaviour rather than treating aggregate robustness as evidence that all
capabilities are preserved equally.

\paragraph{Multilingual and language-aware quantisation:}
Recent work has examined these effects more directly in multilingual settings.
\citet{marchisio2024quantizationmultilingual} show that quantisation
robustness varies substantially across languages and that automatic metrics
can miss degradation visible in generated outputs. Calibration data can also
shape these effects. \citet{zeng-etal-2024-multilingual} show that
multilingual calibration can improve preservation of lower-resource
languages, while \citet{chimoto-etal-2026-calibrating} find that the language
composition of GPTQ and AWQ calibration data affects multilingual
quantisation quality. These studies establish that language-specific
robustness depends on more than bit width alone. However, they primarily
compare distinct languages, where differences in script, typology, content,
and benchmark construction can accompany language identity. This leaves open
whether similar effects arise between a regional variety and its closely
related standard language when the underlying content is held fixed.

\paragraph{Regional and cultural evaluation:}

Accomplishing regional and cultural evaluation requires targeted benchmark datasets that can distinguish
language-specific effects from differences in the evaluated content. This is
particularly challenging for regional and culturally specific language
varieties, which are often unevenly represented in both training data and
existing benchmarks. 
Evaluation of dialectal language skills receives increasing attention in the context of assessing Large language model (LLM) fairness and bias \citep{lin-etal-2025-assessing}.
Often, the focus of these studies is on English, involving synthetically generated data and human-in-the loop validation, e.g.\ by \citet{ziems-etal-2023-multi}, but there are examples involving LLM-as-judge approaches too \citep{faisal-etal-2025-dialectal}.
LLMs have been shown to encode cultural
knowledge unevenly across regions and languages
\citep{santy-etal-2023-nlpositionality, arora-etal-2023-probing}, while
reliably evaluating such knowledge remains an open problem
\citep{hershcovich-etal-2022-challenges}. \citet{li2024culturellmincorporatingculturaldifferences}
propose a method for incorporating cultural differences
into LLMs, while
\citet{durmus2024measuringrepresentationsubjectiveglobal}
investigate how language models represent subjective
opinions across countries.


Recent benchmarks address these concerns through more targeted cultural
evaluation. \citet{chiu2025culturalbench} introduce CulturalBench, comprising
1,696 human-written and human-verified English questions across 45 global
regions, and show that multilingual capability does not necessarily imply
multicultural competence. Closer to our setting,
\citet{john2026bavgroundbenchmarkregionalcultural} introduce \bavground{}, a
benchmark of Bavarian cultural knowledge with aligned Bavarian, Standard
German, and English questions. We use \bavground{} for our quantisation study
and describe the benchmark and its alignment in Section~\ref{sec:setup}.

\section{Experimental Setup}
\label{sec:setup}

We evaluate post-training quantisation by comparing each quantised model with
its own BF16 baseline. The three-way alignment of \bavground{} allows us to
measure performance changes on the same underlying questions in English,
Standard German, and Bavarian. We additionally measure peak GPU memory and
batch-size-one throughput on a single NVIDIA GH200 GPU. Full hardware and
runtime details are provided in Appendix~\ref{tab:experimental-setup-appendix}.

\subsection{Benchmark}
Our work uses the \bavground{} dataset that contains 206 source questions, each realised in English, German
and Bavarian, yielding alltogether 618 multiple-choice instances. The questions of this dataset span eight
cultural domains which are each divided into two subsets: \textsc{gen}, covering broadly accessible (or general)
cultural knowledge, and \textsc{grd}, which consists of question-answer pairs hand-made constructed from regional newspapers,
historical references, and specialist anthropological sources. The evaluated set contains
240 \textsc{gen} and 378 \textsc{grd} instances. Because all three language versions correspond to the same source question,
cross-language comparisons are paired by source item rather than across
independent monolingual benchmarks.

\subsection{Models and Quantisation}

We evaluate 14 open-weight 7B--10B instruction models chosen to cover a range
of language and pretraining profiles. The main panel includes the
Bavarian-adapted \genba{} \citep{hoffmann2025llamagenba10btrilinguallargelanguage}, the German-focused LeoLM and LLaMmlein, five
European multilingual models (EuroLLM, Mistral, Occiglot, Teuken, and
Salamandra), and six broader multilingual models (Llama~3.1, Qwen2.5,
Gemma~2, Aya Expanse, Granite~3.0, and OLMo~2). \genba{} is specifically
adapted to English, German, and Bavarian, while Llama~3.1 provides a closely
related parent-model comparison. Exact checkpoint identifiers are listed in
Appendix~\ref{sec:reproducibility}.

For each model in this main panel, we compare the original BF16 checkpoint
with GPTQ-8, GPTQ-4, and AWQ-4, yielding 56 model--precision configurations.
The GPTQ-8 and GPTQ-4 comparison isolates the effect of precision within the
same quantisation method, while GPTQ-4 and AWQ-4 compare two methods at the
same 4-bit precision. All quantised models use a group size of 128, with
embeddings and output layers retained at their original precision. Quantisation is calibrated on approximately 250k tokens balanced across
English, German, and Bavarian, using translated instruction-style examples
from Alpaca, Dolly, and OpenHermes. These data are used only for post-training quantisation, with no
detected exact-text overlap with \bavground{} questions, answer
options, or recorded source references. 

We additionally test calibration sensitivity on four models by repeating
GPTQ-4 and AWQ-4 with balanced, English-only, German-only, and Bavarian-only
calibration data while keeping the calibration budget, sequence length, and
other quantisation settings fixed (Appendix~\ref{sec:calibration-ablation}). To test whether these effects also depend
on the type of text used for calibration, we repeat the 4-bit experiments
with version-pinned Wikimedia text in English, German, and Bavarian. The calibration-language ablation uses seed 42, while the focused
calibration-source comparison uses three seeds; the latter is additionally
evaluated across all 14 models using balanced seed-42 calibration. Finally, a supplementary Qwen-family analysis evaluates additional
checkpoints from Qwen2, Qwen2.5, and Qwen3 to test whether the behaviour
observed for Qwen2.5 in the main panel extends across Qwen generations and
model sizes.

\subsection{Evaluation Protocol}
\label{sec:evaluation}

Our primary evaluation uses multiple-choice accuracy. For each question, we
select the answer candidate with the highest mean token-level conditional
log-likelihood given the question context. Length normalisation reduces
answer-length effects, while scoring the option text rather than
\texttt{A}--\texttt{D} labels avoids model-specific label priors. We also
evaluate answer-letter likelihood and greedy generation as diagnostics for
label bias and output-format failures, but neither contributes to the primary
accuracy measure. We additionally evaluate free-form generation on five representative models
using all 618 \bavground{} instances. Each quantised response is compared
with its BF16 counterpart by a blinded pairwise LLM judge
(\texttt{gpt-5.6-luna}). Judging prioritises semantic correctness, allows
ties, and considers Bavarian dialect appropriateness secondarily. A random
20\% subset is repeated with reversed response order to check position bias.
Full judge instructions and coverage statistics are given in
Appendix~\ref{sec:freeform-appendix}.

\subsection{Quantisation Effects and Statistical Analysis}

For model $m$, quantisation setting $q$, and language $\ell$, we define the
BF16-relative change as $\Delta_{m,q,\ell}=S_{m,q,\ell}-S_{m,\mathrm{BF16},\ell}$.
We measure whether quantisation widens the pre-existing gap between Bavarian
and comparator language $c$ using

\begin{equation}
A_{m,q,c}
=
\Delta_{m,q,\mathrm{BAR}}
-
\Delta_{m,q,c},
\end{equation}

which we term \emph{dialect-gap amplification}. Negative values indicate that
Bavarian degrades more under quantisation than the comparator language, while
positive values indicate greater degradation of the comparator. We also compare BF16 and quantised predictions item by item to distinguish
correct-to-incorrect, incorrect-to-correct, and correctness-preserving answer
changes. We use 5,000 bootstrap replicates throughout, resampling source
questions for aggregate effects and both models and questions for dialect-gap
amplification. We treat cross-language effects as practically equivalent when
the 90\% confidence interval falls entirely within a $\pm2$ percentage-point
margin, and check the sensitivity of this conclusion at $\pm1$ and $\pm3$
points. Full resampling and sensitivity results are reported in
Appendix~\ref{sec:crossed-bootstrap}.

\section{Results}
\label{sec:results}
We first examine the overall effect of quantisation and whether robustness
differs between Bavarian, its closely related standard language German, and
English. We then analyse how calibration data, prediction instability, and
free-form generation shape this dialect--standard-language comparison,
before considering the resulting deployment trade-offs.

\subsection{Eight-bit robustness is broadly preserved, while four-bit robustness does not follow linguistic specialisation}
\label{sec:results-overall}

\begin{table*}[t]
\centering
\small
\begin{tabular}{lrrrrl}
\toprule
Model & BF16 & GPTQ-8 & GPTQ-4 & AWQ-4 & Best \\
\midrule
Aya Expanse 8B & 58.7 & 58.4 (-0.3) & 53.2 (-5.5) & 58.7 (+0.0) & BF16 \\
EuroLLM 9B & 68.9 & 69.4 (+0.5) & 65.2 (-3.7) & 64.7 (-4.2) & GPTQ-8 \\
GENBA-10B & 57.6 & 57.8 (+0.2) & 51.0 (-6.6) & 56.5 (-1.1) & GPTQ-8 \\
Gemma 2 9B & 55.2 & 55.5 (+0.3) & 55.8 (+0.6) & 58.4 (+3.2) & AWQ-4 \\
Granite 3.0 8B & 62.8 & 62.6 (-0.2) & 62.5 (-0.3) & 57.1 (-5.7) & BF16 \\
LLaMmlein 7B & 57.1 & 57.8 (+0.6) & 57.0 (-0.2) & 58.7 (+1.6) & AWQ-4 \\
LeoLM 7B & 57.0 & 56.6 (-0.3) & 55.2 (-1.8) & 53.4 (-3.6) & BF16 \\
Llama 3.1 8B & 64.7 & 65.0 (+0.3) & 65.2 (+0.5) & 62.8 (-1.9) & GPTQ-4 \\
Mistral 7B v0.3 & 66.8 & 66.3 (-0.5) & 67.2 (+0.3) & 63.6 (-3.2) & GPTQ-4 \\
OLMo 2 7B & 58.7 & 58.7 (+0.0) & 56.0 (-2.8) & 54.5 (-4.2) & BF16 \\
Occiglot 7B & 57.8 & 57.8 (+0.0) & 55.8 (-1.9) & 57.9 (+0.2) & AWQ-4 \\
Qwen2.5 7B & 63.1 & 63.1 (+0.0) & 65.4 (+2.3) & 65.0 (+1.9) & GPTQ-4 \\
Salamandra 7B & 63.4 & 63.4 (+0.0) & 64.2 (+0.8) & 62.6 (-0.8) & GPTQ-4 \\
Teuken 7B & 48.4 & 48.2 (-0.2) & 46.1 (-2.3) & 46.3 (-2.1) & BF16 \\
\bottomrule
\end{tabular}
\caption{
Primary \bavground{} accuracy under post-training quantisation.
Scores are percentages under option-text likelihood evaluation.
Values in parentheses are absolute changes in percentage points relative
to the same model in BF16. The \emph{Best} column denotes the
highest-accuracy precision setting for each model, including BF16.
}
\label{tab:main-results}
\end{table*}

Across the 14 models, GPTQ-8 remains essentially unchanged from BF16, with a
mean accuracy difference of $+0.03$ percentage points. In contrast, both
4-bit methods reduce mean accuracy by about $1.4$--$1.5$ points. The 95\%
confidence interval for GPTQ-8 includes zero, whereas those for GPTQ-4 and
AWQ-4 indicate a clearer aggregate loss. Full confidence intervals are
reported in Appendix~\ref{sec:aggregate-bootstrap}. The average 4-bit loss, however, hides substantial variation across models
(Table~\ref{tab:main-results}). GPTQ-4 effects range from a $2.3$-point gain
for Qwen2.5 to a $6.6$-point loss for \genba{}, while AWQ-4 ranges from a
$3.2$-point gain for Gemma~2 to a $5.7$-point loss for Granite. This variation
does not follow the models' linguistic focus. 

German-oriented models such as
Teuken, LeoLM, and LLaMmlein are not consistently more robust, and the
Bavarian-adapted \genba{} is among the models most affected by GPTQ-4.
Conversely, several broader multilingual models remain comparatively stable.
Apparent gains are checkpoint-dependent rather than family-wide,
as illustrated by the heterogeneous Qwen-family results
(Appendix~\ref{sec:qwen-family}). Taken together, these results suggest that linguistic specialisation does not
consistently protect a model from aggressive compression. GPTQ-8 is stable
across a diverse set of models, whereas 4-bit robustness depends more strongly
on the individual model and quantisation method than on whether the model was
specialised for German, Bavarian, or broader multilingual use.

\subsection{Bavarian's lower baseline does not translate into greater quantisation fragility}
\label{sec:results-language}

\begin{table*}[t]
\centering
\small
\begin{tabular}{lrrrrr}
\toprule
Setting & $\Delta$BAR & $\Delta$DE & $\Delta$EN
        & $A_{\mathrm{BAR-DE}}$ & $A_{\mathrm{BAR-EN}}$ \\
\midrule
GPTQ-8 & -0.1 &  0.0 & +0.2 & -0.1 & -0.3 \\
GPTQ-4 & -1.9 & -0.5 & -1.9 & -1.4 &  0.0 \\
AWQ-4  & -1.1 & -2.1 & -1.1 & +1.0 &  0.0 \\
\bottomrule
\end{tabular}
\caption{Language-specific accuracy changes from BF16, macro-averaged across
the 14 models. Values are percentage points. BAR--DE and BAR--EN compare the
Bavarian change with German and English, respectively; negative values indicate
greater degradation in Bavarian.}
\label{tab:language-amplification}
\end{table*}

These model-level differences also raise the question of whether Bavarian's
weaker baseline makes it more vulnerable to compression. Bavarian starts from
a lower BF16 accuracy than German and English ($56.9\%$ vs.\ $60.7\%$ and
$62.4\%$) and is also tokenised less compactly
(Appendix~\ref{sec:tokenization}). We therefore examine whether quantisation
widens the existing gap between Bavarian and German or English. Table~\ref{tab:language-amplification} shows that this is not the case at
8 bits. The Bavarian--German and Bavarian--English effects both fall within
our $\pm2$ percentage-point equivalence region, indicating that GPTQ-8 largely
preserves the relative language pattern. At 4 bits, however, the effects are
less consistent. GPTQ-4 tends towards greater Bavarian degradation relative
to German, while AWQ-4 shifts the average contrast in the opposite direction;
neither satisfies the equivalence criterion.

Per-model results make this heterogeneity clearer
(Table~\ref{tab:per-model-dialect-effects}). Bavarian accuracy decreases in
10 of 14 models under GPTQ-4 and 8 under AWQ-4, compared with 5 under
GPTQ-8. These losses are not consistently larger than those in German:
under GPTQ-4, Bavarian degrades more than German for 9 models and less for
four, while AWQ-4 is almost evenly split between the two directions. Linguistically specialised models show the same variation. LeoLM, LLaMmlein,
and Teuken lose more in Bavarian under GPTQ-4, whereas \genba{} shows the
opposite pattern, losing $3.4$ points in Bavarian, $5.3$ in German, and
$11.2$ in English. Adaptation to Bavarian therefore does not make a model
uniformly more robust, although it may influence which language is better
preserved under compression. The same disconnect appears beyond language. The harder \textsc{grd} subset
has lower BF16 accuracy than \textsc{gen} ($57.9\%$ vs.\ $63.4\%$), yet both
show nearly identical 4-bit losses
(Appendix~\ref{sec:difficulty-analysis}). Lower baseline performance therefore
does not consistently predict greater quantisation sensitivity. Instead, the
language most affected at 4 bits depends more on the individual model and
quantisation method than on baseline performance or linguistic specialisation.

\subsection{Calibration data reshape 4-bit cross-language robustness}
\label{sec:results-calibration}

The language-specific effects also depend on the calibration corpus
(Table~\ref{tab:calibration-main}). Under GPTQ-4, German-only calibration
produces the strongest Bavarian disadvantage, with a $-3.16$-point
Bavarian--German contrast, whereas Bavarian-only calibration nearly removes
it at $+0.12$ points. Bavarian accuracy itself improves from a $-3.52$-point
change under German-only calibration to only $-0.36$ points under
Bavarian-only calibration. This pattern is consistent across the four models:
German-only calibration always degrades Bavarian more than German, while
Bavarian-only calibration substantially reduces the difference.
\begin{table*}[t]
\centering
\small
\setlength{\tabcolsep}{4.5pt}
\begin{tabular}{lrrrr}
\toprule
& \multicolumn{2}{c}{GPTQ-4} & \multicolumn{2}{c}{AWQ-4} \\
\cmidrule(lr){2-3}\cmidrule(lr){4-5}
Calibration & $\Delta$BAR & BAR--DE & $\Delta$BAR & BAR--DE \\
\midrule
Balanced      & -1.82 & -0.85 & +0.97 & +4.25 \\
English only  & -2.55 & -1.09 & +1.09 & +3.88 \\
German only   & -3.52 & -3.16 & -1.33 & +3.28 \\
Bavarian only & -0.36 & +0.12 & -0.36 & +1.82 \\
\bottomrule
\end{tabular}
\caption{Calibration-language sensitivity in the focused four-model ablation.
Values are mean percentage-point changes from BF16. Negative BAR--DE values
indicate greater Bavarian degradation. Full results are reported in
Appendix~\ref{sec:calibration-ablation}.}
\label{tab:calibration-main}
\end{table*}
AWQ-4 behaves differently. Its mean Bavarian--German contrast remains positive
across all four calibration regimes, indicating greater German than Bavarian
degradation on average. Under German-only calibration, GPTQ-4 and AWQ-4 even
produce opposite Bavarian--German directions for three of the four models.
Calibration composition can therefore change not only the magnitude of the
cross-language effect, but also its direction, with the outcome depending on
the quantisation method.

We also test whether these effects depend on the type of calibration text.
In the focused three-seed comparison, GPTQ-4 remains stable when translated
instruction data are replaced with Wikimedia text. AWQ-4 also changes little
in aggregate, but its Bavarian--German contrast shifts by $-4.63$ points,
showing that calibration source can alter cross-language robustness even when
overall accuracy remains similar. Across finite paired runs in the broader
panel, GPTQ-4 remains comparatively stable, whereas AWQ-4 loses $3.12$
points after excluding one invalid Gemma~2 configuration. Calibration
effects therefore depend on the model and quantisation method as well as
on both the language composition and source of the calibration data
(Appendix~\ref{sec:calibration-source}).

\begin{table*}[t]
\centering
\small
\begin{tabular}{lrrrr}
\toprule
& \multicolumn{2}{c}{Low margin (D1)} &
  \multicolumn{2}{c}{High margin (D10)} \\
\cmidrule(lr){2-3}\cmidrule(lr){4-5}
Setting & Change & Loss & Change & Loss \\
\midrule
GPTQ-8 & 10.16 & 3.23 & 0.00 & 0.00 \\
GPTQ-4 & 39.95 & 15.94 & 1.62 & 0.69 \\
AWQ-4  & 43.42 & 16.28 & 1.27 & 0.23 \\
\bottomrule
\end{tabular}
\caption{Prediction instability by BF16 selected-answer margin. D1 and D10
denote the lowest- and highest-margin deciles, respectively. ``Change'' is
the percentage of paired predictions that differ from BF16, while ``Loss''
is the percentage of all paired rows that are BF16-correct and become
incorrect after quantisation. Mean BF16 margins are 0.025 in D1 and 3.338 in D10.}
\label{tab:margin-main}
\end{table*}

\subsection{Small accuracy losses conceal substantial prediction instability}
\label{sec:results-flips}

Beyond these cross-language differences, aggregate accuracy also hides how
often quantisation changes individual predictions. GPTQ-8 changes only $1.18\%$ of
matched BF16 predictions, compared with $11.93\%$ for GPTQ-4 and $16.00\%$
for AWQ-4. For both 4-bit methods, correct-to-incorrect changes outnumber
compensating gains, with additional wrong-to-wrong changes further increasing
the divergence from BF16. These changes are strongly concentrated on questions where the BF16 model is
already uncertain. GPTQ-4 and AWQ-4 alter roughly $40$--$43\%$ of predictions
in the lowest-margin decile, but only $1$--$2\%$ in the highest; GPTQ-8
changes $10.16\%$ in the lowest decile and none in the highest
(Table~\ref{tab:margin-main}). An item-level analysis confirms that
pre-quantisation decision margin is a much stronger predictor of instability
than language or prompt length (Appendix~\ref{sec:instability-model}).
The small average losses at 4 bits therefore hide substantial changes in individual predictions, particularly on questions where the model was already uncertain before quantisation.

\subsection{Free-form evaluation reveals degradation hidden by multiple-choice accuracy}
\label{sec:results-freeform}

The gap between aggregate accuracy and model behaviour becomes even clearer
in free-form generation. GPTQ-8 remains close to BF16, with $76.1\%$ of
comparisons judged as ties and the quantised response winning or tying in
$86.9\%$ of cases. At 4 bits, BF16 responses are preferred more often,
particularly under AWQ-4 (Table~\ref{tab:freeform-main}).
\begin{table*}[t]
\centering
\small
\setlength{\tabcolsep}{3.5pt}
\begin{tabular}{lrrrrrr}
\toprule
Setting & $N$ & BF16 win & Quant. win & Tie & Not worse & Net quant. \\
\midrule
GPTQ-8 & 3047 & 13.1 & 10.8 & 76.1 & 86.9 & -2.3 \\
GPTQ-4 & 3049 & 36.2 & 29.8 & 33.9 & 63.7 & -6.4 \\
AWQ-4  & 3048 & 45.7 & 28.1 & 26.1 & 54.2 & -17.6 \\
\bottomrule
\end{tabular}
\caption{Blinded pairwise comparison of BF16 and quantised free-form responses
across five models and all \bavground{} instances. $N$ is the number of
successfully parsed comparisons. ``Not worse'' combines quantised wins and
ties; net preference is the quantised-win rate minus the BF16-win rate.}
\label{tab:freeform-main}
\end{table*}
The difference between the two 4-bit methods is much larger here than in
multiple-choice accuracy. GPTQ-4 and AWQ-4 have similar average
multiple-choice losses, yet AWQ-4 shows substantially poorer retention in
free-form responses. This gap is also visible at the model level. For example,
Qwen2.5~7B improves on the multiple-choice benchmark under GPTQ-4, while its
BF16 responses are still preferred overall in the free-form comparison.
This shows that a small gain in fixed-choice accuracy does not necessarily
carry over to generated responses. 

The same mismatch between fixed-choice accuracy and generated responses is
also visible across languages. GPTQ-8 remains close to BF16 in Bavarian,
German, and English. At 4 bits, however, the degradation is more pronounced.
Under GPTQ-4, BF16 responses are preferred more strongly in Bavarian and
German, while English remains comparatively stable. AWQ-4 shows degradation
across all three languages, with German at least as affected as Bavarian.
This mirrors the multiple-choice results above: 4-bit compression can affect
Bavarian substantially, but the effect is not consistently stronger than in
German. Because the free-form comparison relies on pairwise judging, we also checked sensitivity to response order, which was generally consistent. We treat these results as complementary evidence of response preservation rather than as human validation of Bavarian naturalness or sociolinguistic quality (Appendix~\ref{sec:freeform-appendix}).

\subsection{Large memory savings do not require proportional quality loss}
\label{sec:results-efficiency}

\begin{figure}[t]
    \centering
    \includegraphics[height=0.2\textheight,keepaspectratio]{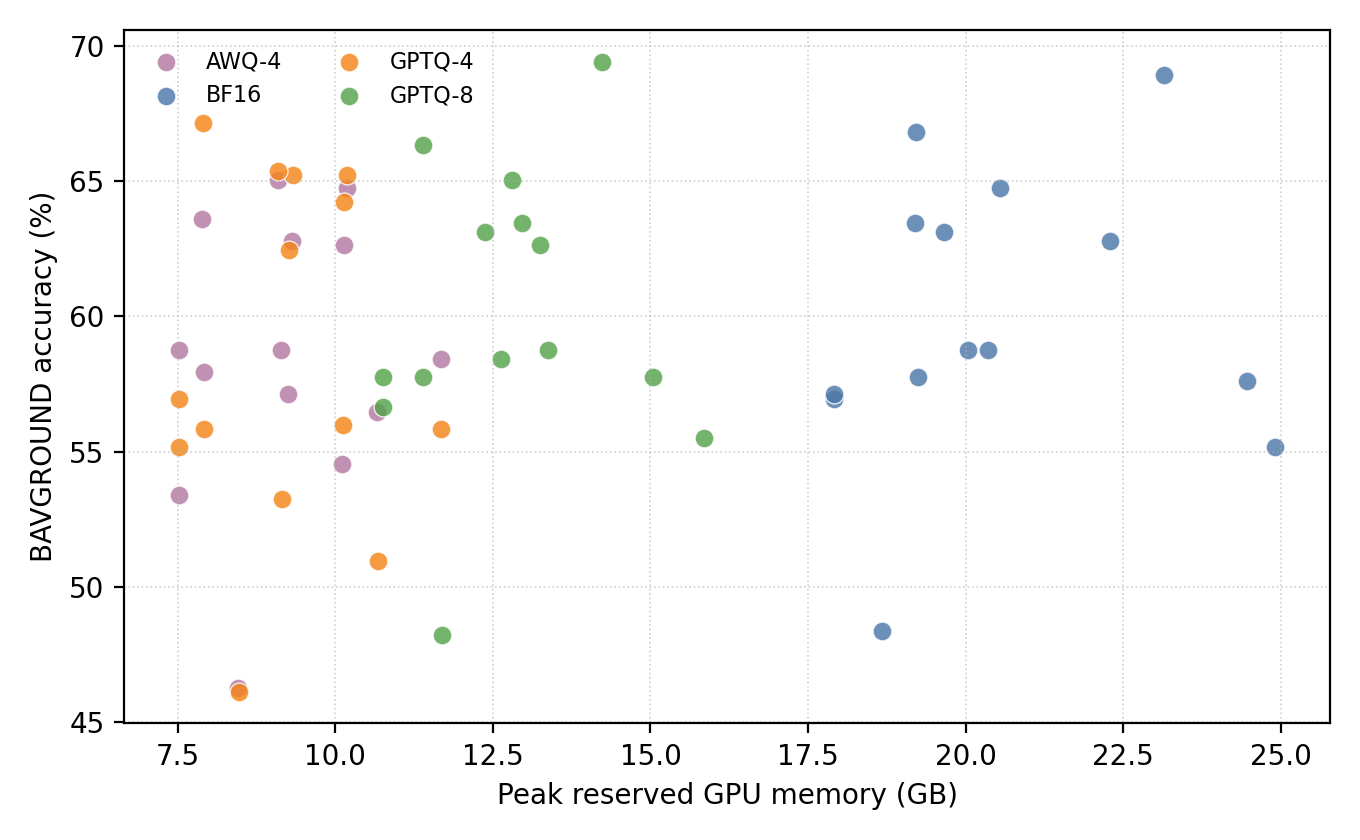}
    \caption{Quality--memory frontier across completed model and precision
    settings. Points show \bavground{} multiple-choice accuracy against peak
    reserved GPU memory.}
    \label{fig:pareto}
\end{figure}

Despite the quality losses observed above, quantisation can still offer
substantial memory savings with little change in multiple-choice accuracy.
Across the measured configurations, 11 quantised model configurations reduce peak reserved memory by at least
$50\%$ while losing no more than one percentage point relative to BF16
(Figure~\ref{fig:pareto}). Mistral~7B GPTQ-4 provides one of the strongest
quality--memory trade-offs in our measurements: memory falls from
$19.2$\,GB to $7.9$\,GB, a $58.9\%$ reduction, while multiple-choice accuracy
remains essentially unchanged ($66.8\%$ vs.\ $67.2\%$)
(Appendix~\ref{sec:efficiency-appendix}). Its Bavarian, German, and English
multiple-choice scores also remain close to their BF16 baselines
(Appendix~\ref{sec:per-model-dialect-effects}). The same Mistral configuration also shows why this trade-off depends on how
quality is measured. Although its multiple-choice accuracy is preserved,
free-form judging gives GPTQ-4 a $-3.9$ point net preference relative to BF16
overall and $-5.1$ points in Bavarian
(Appendix~\ref{sec:freeform-appendix}). A strong quality--memory trade-off
under fixed-choice evaluation can therefore still hide degradation in
generated responses. More broadly, GPTQ-4 matches or exceeds GPTQ-8
multiple-choice accuracy for 5 of 14 models despite using fewer bits,
showing that greater compression is not uniformly worse.

\section{Discussion}

The clearest distinction in our results is between broadly stable 8-bit
quantisation and less predictable 4-bit compression. GPTQ-8 largely
preserves average accuracy and the relative behaviour of Bavarian, German,
and English. At 4 bits, however, robustness varies more across models and
methods, with no consistent protection from linguistic specialisation. This
makes calibration particularly important. 

Changing the calibration language can alter the Bavarian--German effect under
GPTQ-4, while calibration source reveals a different method-dependent pattern.
GPTQ-4 remains comparatively stable across instruction-style and Wikimedia
data, whereas AWQ-4 shows stronger cross-language shifts and model-specific
failures. Robustness therefore depends not only on the checkpoint and method,
but also on the calibration data. These differences are also difficult to explain from baseline performance.
Bavarian starts from lower BF16 accuracy and less compact tokenisation, yet
neither consistently predicts larger losses. Instead, prediction changes
concentrate on low-margin BF16 decisions, explaining why small average
accuracy changes can conceal substantial behavioural differences. 

The
free-form results show the same mismatch: GPTQ-4 and AWQ-4 have similar
fixed-choice losses, but AWQ-4 preserves generated responses less well. For closely related varieties, the question is therefore not only whether
quantisation lowers performance, but whether it changes the relative behaviour
of the dialect and standard language. In our aligned Bavarian--German
comparison, this depends on the model, quantisation method, calibration data,
and evaluation setting rather than on baseline performance or linguistic
specialisation alone. Robustness should therefore be assessed via
language-specific changes on matched questions, not
aggregate multilingual accuracy.

\section{Conclusion}

We asked whether making language models lighter through post-training
quantisation also preserves their behaviour across Bavarian, Standard German,
and English. For GPTQ-8, the answer is largely yes. Performance remains close
to BF16 and the relative differences between the three languages are broadly
preserved. At 4 bits, this stability weakens. Bavarian is not consistently
more affected than German, but the relative behaviour of the two languages
varies more across models, quantisation methods, and calibration data. The aligned dialect vs.\ standard language comparison is important here because these
changes are difficult to identify from aggregate multilingual accuracy
alone. It also shows that lower baseline performance in Bavarian does not by
itself imply greater fragility, while free-form evaluation reveals losses that
fixed-choice accuracy can miss. Our matrix of experiments indicates that quantisation can therefore make models
substantially lighter without necessarily harming Bavarian more than German,
but linguistic preservation at 4 bits cannot be assumed from compression
level alone. For regional language varieties, lightweight deployment and
linguistic robustness need to be evaluated together.

\section{Limitations}

Our study focuses on Bavarian and \bavground{}, so the findings may not
generalise to other dialects or low-resource languages, and we do not model
Bavarian's internal regional or orthographic variation. The free-form analysis broadens the evaluation beyond multiple choice but relies on a single LLM judge. The response-order audit shows high consistency, out of 9,270 planned
comparisons, 112 were excluded due to output truncation and 14 due
to malformed pair identifiers. The calibration-language ablation is deliberately focused, covering four
models and one seed per condition. The calibration-source experiment adds three-seed replication for the
focused four-model comparison while its broader seed-42 comparison uses
balanced calibration, with 14 valid GPTQ-4 and 13 valid AWQ-4 pairs.
Broader replication would therefore be useful for establishing how stable
these calibration effects are across models and seeds. Finally, the memory results are specific to our hardware, software stack, and
batch-size-one inference setup.

\paragraph{AI Assistance Usage Statement:}
Generative AI tools were used for language editing, formatting assistance,
and support in generating figures. All resulting text, figures, references,
and numerical values were reviewed and verified.

\bibliography{emnlp26}

\appendix


\section{Additional Analyses}
\label{sec:additional-analysis}

\subsection{Aggregate Bootstrap Confidence Intervals}
\label{sec:aggregate-bootstrap}

Table~\ref{tab:aggregate-bootstrap} reports the clustered-bootstrap
confidence intervals for the aggregate BF16-relative accuracy differences
used in Section~\ref{sec:results-overall}. We resample the 206 aligned source
questions and retain all associated English, German, and Bavarian instances
for each sampled source question. These intervals therefore quantify item
uncertainty conditional on the evaluated 14-model panel.

\begin{table*}[t]
\centering
\small
\begin{tabular}{lrrrr}
\toprule
Setting & $\Delta$ vs.\ BF16 & 95\% CI & Source clusters & Paired rows \\
\midrule
GPTQ-8 & +0.03 & [-0.14, +0.21] & 206 & 8,652 \\
GPTQ-4 & -1.47 & [-2.21, -0.72] & 206 & 8,652 \\
AWQ-4  & -1.42 & [-2.21, -0.61] & 206 & 8,652 \\
\bottomrule
\end{tabular}
\caption{
Aggregate accuracy change under quantisation, measured in percentage points
relative to each model's BF16 baseline. Confidence intervals are 95\%
percentile intervals from 5,000 bootstrap samples clustered by
source question.
}
\label{tab:aggregate-bootstrap}
\end{table*}

\subsection{Focused Qwen-Family Analysis}
\label{sec:qwen-family}

To probe whether the Qwen2.5~7B improvement in the main matrix is representative
of the broader family, we additionally evaluate five Qwen checkpoints.
Table~\ref{tab:qwen-family} summarises these results. This is a descriptive
family analysis rather than a clean scaling study because model generation and
architecture vary across several rows.

\begin{table*}[t]
\centering
\small
\begin{tabular}{lrrr}
\toprule
Model & BF16 & Best quantised & $\Delta$ \\
\midrule
Qwen2 7B Instruct & 63.8 & 63.6 (GPTQ-8) & -0.2 \\
Qwen2.5 3B Instruct & 60.7 & 60.4 (GPTQ-8) & -0.3 \\
Qwen2.5 7B Instruct & 63.1 & 65.4 (GPTQ-4) & +2.3 \\
Qwen3 4B & 51.9 & 51.6 (GPTQ-8) & -0.3 \\
Qwen3 8B & 48.1 & 56.6 (GPTQ-4) & +8.6 \\
\bottomrule
\end{tabular}
\caption{Focused Qwen-family results on \bavground{}. Scores are overall
multiple-choice accuracy (\%). ``Best quantised'' selects the highest observed
quantised configuration for each checkpoint.}
\label{tab:qwen-family}
\end{table*}

As Table~\ref{tab:qwen-family} shows, the Qwen2.5~7B gain is not representative
of all checkpoints in the family: Qwen2~7B, Qwen2.5~3B, and Qwen3~4B are best
preserved by GPTQ-8, while Qwen2.5~7B and Qwen3~8B show their highest
multiple-choice scores under GPTQ-4. These results therefore indicate
substantial checkpoint-level heterogeneity rather than a consistent family-wide
benefit from 4-bit quantisation.

Within Qwen2.5, the 7B checkpoint is $2.4$ points stronger than the 3B
checkpoint in BF16 overall and $6.8$ points stronger on Bavarian. Across model
generations, however, parameter count does not predict either baseline quality
or quantisation response. In particular, Qwen3~8B shows a large $+8.6$-point
GPTQ-4 increase from a comparatively weak BF16 baseline. We report this result
descriptively and do not treat it as evidence that 4-bit quantisation improves
generation quality; the free-form results for Qwen2.5 already demonstrate that
multiple-choice gains can coexist with worse pairwise output preferences.

\subsection{Crossed Bootstrap and Practical Equivalence}
\label{sec:crossed-bootstrap}

Table~\ref{tab:crossed-equivalence} reports crossed-bootstrap estimates and
confidence intervals for dialect-gap amplification. We jointly resample models
and source questions to capture uncertainty from both the evaluated model set
and the 206 aligned \bavground{} source questions. We use 5,000 replicates and
report 90\% and 95\% confidence intervals.

Because a non-significant difference does not establish comparable robustness,
we also assess practical equivalence using a $\pm2$ percentage-point margin.
Two language effects are considered practically equivalent only when the full
90\% confidence interval lies within this region.

\begin{table*}[t]
\centering
\small
\begin{tabular}{llrrrl}
\toprule
Setting & Contrast & Estimate & 90\% CI & 95\% CI & Equivalent \\
\midrule
GPTQ-8 & BAR--DE & -0.07 & [-0.73, +0.59] & [-0.83, +0.69] & Yes \\
GPTQ-8 & BAR--EN & -0.35 & [-1.04, +0.31] & [-1.21, +0.45] & Yes \\
GPTQ-4 & BAR--DE & -1.42 & [-3.43, +0.59] & [-3.81, +0.90] & No \\
GPTQ-4 & BAR--EN & +0.00 & [-2.08, +2.18] & [-2.46, +2.67] & No \\
AWQ-4  & BAR--DE & +1.01 & [-1.35, +3.40] & [-1.94, +3.88] & No \\
AWQ-4  & BAR--EN & +0.03 & [-2.32, +2.50] & [-2.77, +3.02] & No \\
\bottomrule
\end{tabular}
\caption{
Dialect-gap amplification under a crossed bootstrap over models and source
questions. Values are percentage points. Negative estimates indicate greater
Bavarian degradation. Practical equivalence requires the 90\% confidence
interval to lie entirely within the $\pm2$ percentage-point region.
}
\label{tab:crossed-equivalence}
\end{table*}

GPTQ-8 is the only setting for which both Bavarian--German and
Bavarian--English contrasts satisfy the equivalence criterion. Neither 4-bit
method establishes equivalence. The Bavarian--German point estimates also
differ in direction between GPTQ-4 and AWQ-4, reinforcing that the 4-bit
language effect is not consistent across quantisation methods.

We additionally test the sensitivity of these conclusions to $\pm1$, $\pm2$,
and $\pm3$ percentage-point margins (Table~\ref{tab:sesoi-sensitivity}).
GPTQ-8 remains the most robust setting: its Bavarian--German contrast is
equivalent even under the stricter $\pm1$-point margin, while its
Bavarian--English contrast is equivalent from $\pm2$ points onward. Neither
4-bit Bavarian--German contrast reaches equivalence even at $\pm3$ points,
whereas the Bavarian--English contrasts for GPTQ-4 and AWQ-4 become equivalent
only under this wider margin.

\begin{table*}[t]
\centering
\small
\begin{tabular}{llccc}
\toprule
Setting & Contrast & $\pm1$ pp & $\pm2$ pp & $\pm3$ pp \\
\midrule
GPTQ-8 & BAR--DE & Yes & Yes & Yes \\
GPTQ-8 & BAR--EN & No  & Yes & Yes \\
GPTQ-4 & BAR--DE & No  & No  & No \\
GPTQ-4 & BAR--EN & No  & No  & Yes \\
AWQ-4  & BAR--DE & No  & No  & No \\
AWQ-4  & BAR--EN & No  & No  & Yes \\
\bottomrule
\end{tabular}
\caption{
Sensitivity of practical-equivalence conclusions to alternative equivalence
margins. Entries indicate whether the 90\% crossed-bootstrap confidence
interval lies fully within the stated percentage-point region.
}
\label{tab:sesoi-sensitivity}
\end{table*}

\subsection{Per-Model Language Effects}
\label{sec:per-model-dialect-effects}

The aggregate dialect-gap estimates average over substantial model-level
variation. We therefore examine the direction of the Bavarian--German effect
for each model before reporting the full per-model results. Table~\ref{tab:per-model-dialect-signs}
summarises how often Bavarian degrades more than German, how often the two
effects are similar, and how often German degrades more than Bavarian.

\begin{table*}[t]
\centering
\small
\begin{tabular}{lrrr}
\toprule
Setting & BAR $<$ DE & Similar & BAR $>$ DE \\
\midrule
GPTQ-8 & 6 & 2 & 6 \\
GPTQ-4 & 9 & 1 & 4 \\
AWQ-4  & 6 & 1 & 7 \\
\bottomrule
\end{tabular}
\caption{
Direction of the per-model Bavarian--German quantisation effect. BAR $<$ DE
indicates greater Bavarian degradation, while BAR $>$ DE indicates greater
German degradation. Values within $\pm0.25$ percentage points are grouped as
similar.
}
\label{tab:per-model-dialect-signs}
\end{table*}

The sign pattern reinforces the aggregate results. GPTQ-8 is nearly balanced
across directions and the individual contrasts are generally small. GPTQ-4
shows a clearer tendency towards greater Bavarian degradation, with 9 of 14
models exhibiting a negative Bavarian--German contrast, although the effect is
not universal. AWQ-4 is more evenly split, with greater Bavarian degradation
for six models and greater German degradation for seven. Thus, the direction of
the 4-bit language effect remains strongly model- and quantiser-dependent.

Table~\ref{tab:per-model-dialect-effects} reports the corresponding
language-specific changes from BF16 for every model and quantisation setting.
The Bavarian- and German-focused models are also heterogeneous. LeoLM,
LLaMmlein, and Teuken show larger Bavarian than German losses under GPTQ-4,
whereas \genba{} shows the opposite pattern, with its largest GPTQ-4
degradation occurring in English. Several of these directions change again
under AWQ-4, providing no evidence that linguistic specialisation alone
determines language-specific robustness under quantisation.
\begin{table*}[t]
\centering
\scriptsize
\setlength{\tabcolsep}{3pt}
\begin{tabular}{llrrrrr}
\toprule
\textbf{Model} & \textbf{Quantiser} & \textbf{$\Delta$BAR} & \textbf{$\Delta$DE} & \textbf{$\Delta$EN} & \textbf{BAR-DE} & \textbf{BAR-EN} \\
\midrule
Aya Expanse 8B & GPTQ-8 & +0.0 & -0.5 & -0.5 & +0.5 & +0.5 \\
Aya Expanse 8B & GPTQ-4 & -7.8 & +0.0 & -8.7 & -7.8 & +1.0 \\
Aya Expanse 8B & AWQ-4 & +0.0 & +2.4 & -2.4 & -2.4 & +2.4 \\
EuroLLM 9B & GPTQ-8 & +1.0 & +0.5 & +0.0 & +0.5 & +1.0 \\
EuroLLM 9B & GPTQ-4 & -4.9 & -1.9 & -4.4 & -2.9 & -0.5 \\
EuroLLM 9B & AWQ-4 & -1.5 & -8.3 & -2.9 & +6.8 & +1.5 \\
GENBA-10B & GPTQ-8 & +0.0 & +0.0 & +0.5 & +0.0 & -0.5 \\
GENBA-10B & GPTQ-4 & -3.4 & -5.3 & -11.2 & +1.9 & +7.8 \\
GENBA-10B & AWQ-4 & +2.9 & -3.4 & -2.9 & +6.3 & +5.8 \\
Gemma 2 9B & GPTQ-8 & +0.5 & +0.0 & +0.5 & +0.5 & +0.0 \\
Gemma 2 9B & GPTQ-4 & -0.5 & +2.4 & +0.0 & -2.9 & -0.5 \\
Gemma 2 9B & AWQ-4 & +0.0 & +5.3 & +4.4 & -5.3 & -4.4 \\
Granite 3.0 8B & GPTQ-8 & +0.0 & -0.5 & +0.0 & +0.5 & +0.0 \\
Granite 3.0 8B & GPTQ-4 & +0.5 & -0.5 & -1.0 & +1.0 & +1.5 \\
Granite 3.0 8B & AWQ-4 & -3.4 & -8.7 & -4.9 & +5.3 & +1.5 \\
LLaMmlein 7B & GPTQ-8 & -0.5 & +1.0 & +1.5 & -1.5 & -1.9 \\
LLaMmlein 7B & GPTQ-4 & -1.9 & +0.5 & +1.0 & -2.4 & -2.9 \\
LLaMmlein 7B & AWQ-4 & +1.0 & -1.0 & +4.9 & +1.9 & -3.9 \\
LeoLM 7B & GPTQ-8 & -1.0 & -0.5 & +0.5 & -0.5 & -1.5 \\
LeoLM 7B & GPTQ-4 & -3.4 & +1.0 & -2.9 & -4.4 & -0.5 \\
LeoLM 7B & AWQ-4 & -4.4 & -0.5 & -5.8 & -3.9 & +1.5 \\
Llama 3.1 8B & GPTQ-8 & +0.0 & +0.5 & +0.5 & -0.5 & -0.5 \\
Llama 3.1 8B & GPTQ-4 & -1.0 & -0.5 & +2.9 & -0.5 & -3.9 \\
Llama 3.1 8B & AWQ-4 & -0.5 & -0.5 & -4.9 & +0.0 & +4.4 \\
Mistral 7B v0.3 & GPTQ-8 & -1.5 & -1.0 & +1.0 & -0.5 & -2.4 \\
Mistral 7B v0.3 & GPTQ-4 & +0.5 & +0.5 & +0.0 & +0.0 & +0.5 \\
Mistral 7B v0.3 & AWQ-4 & -4.9 & -3.9 & -1.0 & -1.0 & -3.9 \\
OLMo 2 7B & GPTQ-8 & +0.5 & -0.5 & +0.0 & +1.0 & +0.5 \\
OLMo 2 7B & GPTQ-4 & -2.4 & -3.9 & -1.9 & +1.5 & -0.5 \\
OLMo 2 7B & AWQ-4 & -3.4 & -8.3 & -1.0 & +4.9 & -2.4 \\
Occiglot 7B & GPTQ-8 & +0.5 & +0.0 & -0.5 & +0.5 & +1.0 \\
Occiglot 7B & GPTQ-4 & -2.9 & -1.5 & -1.5 & -1.5 & -1.5 \\
Occiglot 7B & AWQ-4 & +1.0 & -0.5 & +0.0 & +1.5 & +1.0 \\
Qwen2.5 7B & GPTQ-8 & -0.5 & +0.5 & +0.0 & -1.0 & -0.5 \\
Qwen2.5 7B & GPTQ-4 & +1.9 & +3.9 & +1.0 & -1.9 & +1.0 \\
Qwen2.5 7B & AWQ-4 & +2.9 & -1.0 & +3.9 & +3.9 & -1.0 \\
Salamandra 7B & GPTQ-8 & +0.0 & +0.0 & +0.0 & +0.0 & +0.0 \\
Salamandra 7B & GPTQ-4 & +1.9 & +0.5 & +0.0 & +1.5 & +1.9 \\
Salamandra 7B & AWQ-4 & -0.5 & +0.5 & -2.4 & -1.0 & +1.9 \\
Teuken 7B & GPTQ-8 & -0.5 & +0.0 & +0.0 & -0.5 & -0.5 \\
Teuken 7B & GPTQ-4 & -3.9 & -2.4 & -0.5 & -1.5 & -3.4 \\
Teuken 7B & AWQ-4 & -4.4 & -1.5 & -0.5 & -2.9 & -3.9 \\
\bottomrule
\end{tabular}
\caption{
Per-model language-specific accuracy changes under balanced trilingual
calibration. $\Delta$BAR, $\Delta$DE, and $\Delta$EN denote percentage-point
changes from each model's BF16 accuracy. BAR--DE and BAR--EN compare these
changes across languages; negative values indicate greater Bavarian
degradation.
}
\label{tab:per-model-dialect-effects}
\end{table*}

\subsection{Tokenizer Segmentation}
\label{sec:tokenization}

\begin{table*}[t]
\centering
\scriptsize
\begin{tabular}{lrrrrr}
\toprule
Domain &
Prompt BAR/DE &
Prompt BAR/EN &
Prompt DE/EN &
Gold BAR/DE &
Gold BAR/EN \\
\midrule
Building \& Sacred Heritage & 1.04 & 1.18 & 1.14 & 1.09 & 1.30 \\
Language & 1.00 & 1.10 & 1.09 & 0.94 & 1.06 \\
Politics & 1.03 & 1.17 & 1.14 & 1.01 & 1.31 \\
Historical & 1.05 & 1.17 & 1.12 & 1.07 & 1.20 \\
Landscape & 1.03 & 1.14 & 1.11 & 1.05 & 1.22 \\
Living Traditions \& Customs & 1.03 & 1.18 & 1.15 & 1.03 & 1.38 \\
Arts \& Identity & 1.03 & 1.18 & 1.15 & 1.06 & 1.26 \\
Culinary & 1.02 & 1.20 & 1.17 & 1.01 & 1.43 \\
\bottomrule
\end{tabular}
\caption{
Relative tokenisation length across aligned language versions, computed from
BF16 runs and paired by model and source question. Ratios above 1 indicate
that the numerator language requires more tokens. Gold denotes the gold answer
option.
}
\label{tab:tokenizer-ratios}
\end{table*}

Bavarian prompts are generally tokenised slightly less compactly than German
and substantially less compactly than English as seen in Table~\ref{tab:tokenizer-ratios}. Across domains, BAR/DE prompt
ratios range from $1.00$ to $1.05$, while BAR/EN ratios range from $1.10$ to
$1.20$. Gold-answer ratios show a similar tendency, although the magnitude
varies by domain. Since Bavarian does not consistently exhibit greater
quantisation degradation, tokenisation length alone is insufficient to explain
the observed language-level robustness differences.

\subsection{Task difficulty and quantisation sensitivity}
\label{sec:difficulty-analysis}

\begin{table*}[t]
\centering
\small
\begin{tabular}{lrrrr}
\toprule
Subset & BF16 & $\Delta$GPTQ-8 & $\Delta$GPTQ-4 & $\Delta$AWQ-4 \\
\midrule
GEN & 63.4 & +0.2 & -1.4 & -1.4 \\
GRD & 57.9 & -0.1 & -1.5 & -1.4 \\
\bottomrule
\end{tabular}
\caption{
Performance on general (\textsc{gen}) and source-grounded
(\textsc{grd}) \bavground{} questions, macro-averaged across the
14 models. BF16 reports absolute option-text accuracy (\%);
quantised columns report the mean absolute change from each model's
BF16 score in percentage points.
}
\label{tab:subset-results}
\end{table*}

Table~\ref{tab:subset-results} gives the full comparison between the
\textsc{gen} and \textsc{grd} subsets. Source-grounded questions are harder
before quantisation, with a mean BF16 accuracy of $57.9\%$, compared with
$63.4\%$ for general cultural questions. The difference in baseline
difficulty is not reflected in their response to compression. GPTQ-4 reduces
accuracy by $1.5$ points on \textsc{grd} and $1.4$ on \textsc{gen}, while
AWQ-4 reduces both by approximately $1.4$ points. GPTQ-8 leaves both subsets
essentially unchanged.

The same lack of correspondence appears across cultural domains. Language is
the hardest domain in BF16, at $40.5\%$, and loses $2.2$ points under GPTQ-4
and $1.9$ under AWQ-4. Historical, despite having the highest BF16 accuracy at
$68.3\%$, shows similarly large losses of $2.2$ and $3.0$ points. Politics,
in contrast, reaches $61.7\%$ in BF16 but changes little under either 4-bit
method ($+0.4$ for GPTQ-4 and $-0.1$ for AWQ-4). Across both question types
and domains, lower baseline accuracy therefore does not consistently
correspond to greater quantisation sensitivity. Full domain-level results are
reported in Appendix~\ref{sec:domain-analysis}.


\subsection{Domain-Level Quantisation Effects}
\label{sec:domain-analysis}

\begin{table*}[t]
\centering
\small
\begin{tabular}{lrrrr}
\toprule
Domain & BF16 & $\Delta$GPTQ-8 & $\Delta$GPTQ-4 & $\Delta$AWQ-4 \\
\midrule
Language & 40.5 & -0.4 & -2.2 & -1.9 \\
Building \& Sacred Heritage & 57.1 & +0.0 & -2.4 & -3.1 \\
Politics & 61.7 & -0.3 & +0.4 & -0.1 \\
Arts \& Identity & 62.7 & +0.2 & -1.6 & -0.9 \\
Culinary & 62.8 & +0.3 & +0.0 & +0.1 \\
Landscape & 63.4 & +0.3 & -2.4 & -1.6 \\
Living Traditions \& Customs & 63.9 & +0.2 & -1.2 & -0.7 \\
Historical & 68.3 & -0.0 & -2.2 & -3.0 \\
\bottomrule
\end{tabular}
\caption{
Domain-level \bavground{} performance averaged across the 14 models.
BF16 reports absolute option-text accuracy (\%); quantised columns report
absolute percentage-point changes from each model's BF16 score before averaging.
}
\label{tab:domain-results}
\end{table*}

The domain-level results show that baseline difficulty does not monotonically
predict quantisation sensitivity. Language is the weakest BF16 domain
($40.5\%$) and loses $2.2$ and $1.9$ points under GPTQ-4 and AWQ-4,
respectively. However, Historical is the strongest BF16 domain ($68.3\%$)
and experiences similarly large drops of $2.2$ and $3.0$ points. Politics,
by contrast, remains almost unchanged under both 4-bit settings despite
intermediate BF16 performance.

\begin{table*}[t]
\centering
\small
\setlength{\tabcolsep}{4pt}
\begin{tabular}{ll}
\toprule
\textbf{Model} & \textbf{Checkpoint identifier} \\
\midrule
GENBA-10B & \texttt{GENBA-10B-instruct \cite{hoffmann2025llamagenba10btrilinguallargelanguage}} \\
Llama 3.1 Instruct & \texttt{meta-llama/Llama-3.1-8B-Instruct} \\
EuroLLM Instruct & \texttt{utter-project/EuroLLM-9B-Instruct} \\
Qwen2.5 Instruct & \texttt{Qwen/Qwen2.5-7B-Instruct} \\
Mistral Instruct v0.3 & \texttt{mistralai/Mistral-7B-Instruct-v0.3} \\
Gemma 2 IT & \texttt{google/gemma-2-9b-it} \\
Aya Expanse & \texttt{CohereLabs/aya-expanse-8b} \\
Granite 3.0 Instruct & \texttt{ibm-granite/granite-3.0-8b-instruct} \\
LeoLM Chat & \texttt{LeoLM/leo-hessianai-7b-chat} \\
LLaMmlein Chat & \texttt{LSX-UniWue/LLaMmlein\_7B\_chat} \\
Occiglot EU5 Instruct & \texttt{occiglot/occiglot-7b-eu5-instruct} \\
Teuken Instruct & \texttt{openGPT-X/Teuken-7B-instruct-research-v0.4} \\
Salamandra Instruct & \texttt{BSC-LT/salamandra-7b-instruct} \\
OLMo 2 Instruct & \texttt{allenai/OLMo-2-1124-7B-Instruct} \\
\bottomrule
\end{tabular}
\caption{Exact checkpoint identifiers used in the quantisation study.}
\label{tab:checkpoint-identifiers}
\end{table*}

\subsection{Calibration-Language Ablation}
\label{sec:calibration-ablation}

\paragraph{Models and calibration corpus construction.}
The focused calibration-language ablation uses four instruction-tuned
models: GENBA-10B, Llama~3.1~8B, Qwen2.5~7B, and EuroLLM~9B.
Exact checkpoint identifiers are provided in
Table~\ref{tab:checkpoint-identifiers}.
Calibration data are drawn from language-specific English, German,
and Bavarian instruction corpora derived from Alpaca, Dolly, and
OpenHermes. Each language pool is independently shuffled using
Python seed 42, and complete records are selected until the
target budget is reached. The balanced condition allocates
approximately 83,333 whitespace-delimited tokens per language,
whereas each monolingual condition uses approximately 250,000
tokens from a single language. Budgets are measured using
whitespace tokenisation rather than model-specific tokenisers.
The selected examples are not aligned translations, and their
source mixtures may differ across languages. Consequently,
language is confounded with corpus content, topic, and style;
we interpret these results as an exploratory calibration-corpus
composition ablation rather than an isolated causal effect
of calibration language. Exact corpus sizes are reported in
Table~\ref{tab:calibration-corpus-statistics}.

\begin{table*}[t]
\centering
\small
\setlength{\tabcolsep}{5pt}
\begin{tabular}{lrrrr}
\toprule
Calibration & Documents & EN tokens & DE tokens & BAR tokens \\
\midrule
Balanced & 1,383 & 83,416 & 83,423 & 83,607 \\
EN-only  & 1,369 & 250,238 & -- & -- \\
DE-only  & 1,387 & -- & 250,291 & -- \\
BAR-only & 1,363 & -- & -- & 250,106 \\
\bottomrule
\end{tabular}
\caption{Calibration corpus statistics for seed 42.
Token counts use whitespace-delimited tokens rather
than model-specific tokenisers. Language pools are
sampled independently, without aligned source-example IDs.
Exact normalised-text checks found no overlap with
\bavground{} questions, answer options, or recorded source references.}
\label{tab:calibration-corpus-statistics}
\end{table*}

Table~\ref{tab:calibration-full} expands the calibration analysis from
Section~\ref{sec:results-calibration}. Values are macro-averaged across the
four models in the focused ablation and are measured relative to each model's
own BF16 checkpoint.

\begin{table*}[t]
\centering
\small
\begin{tabular}{llrrrrr}
\toprule
Setting & Calibration & $\Delta$EN & $\Delta$DE & $\Delta$BAR & BAR--DE & BAR--EN \\
\midrule
GPTQ-4 & Balanced & -2.91 & -0.97 & -1.82 & -0.85 & +1.09 \\
GPTQ-4 & English only & -2.79 & -1.46 & -2.55 & -1.09 & +0.24 \\
GPTQ-4 & German only & -0.12 & -0.36 & -3.52 & -3.16 & -3.40 \\
GPTQ-4 & Bavarian only & -0.85 & -0.49 & -0.36 & +0.12 & +0.49 \\
\midrule
AWQ-4 & Balanced & -1.70 & -3.28 & +0.97 & +4.25 & +2.67 \\
AWQ-4 & English only & -1.33 & -2.79 & +1.09 & +3.88 & +2.43 \\
AWQ-4 & German only & -2.55 & -4.61 & -1.33 & +3.28 & +1.21 \\
AWQ-4 & Bavarian only & -1.70 & -2.18 & -0.36 & +1.82 & +1.33 \\
\bottomrule
\end{tabular}
\caption{Full calibration-language ablation. All values are mean accuracy
changes in percentage points relative to BF16. Positive BAR--DE indicates that
German degrades more than Bavarian; negative values indicate greater Bavarian
degradation.}
\label{tab:calibration-full}
\end{table*}

As Table~\ref{tab:calibration-full} shows, calibration composition can
substantially change the cross-language effect of 4-bit quantisation.
For GPTQ-4, German-only calibration produces the strongest Bavarian
disadvantage, whereas Bavarian-only calibration nearly removes the
Bavarian--German differential. AWQ-4 shows the opposite aggregate direction
across all four regimes, with German generally degrading more than Bavarian,
although the magnitude varies with calibration language.

To assess whether these averages are driven by individual models,
Table~\ref{tab:calibration-signs} summarises the direction of the
Bavarian--German effect, with the individual estimates reported in
Table~\ref{tab:calibration-per-model}. The German-only GPTQ-4 result is the most consistent case:
Bavarian degrades more than German for all four models, with BAR--DE ranging
from $-6.8$ to $-0.5$ points. Under Bavarian-only calibration, the GPTQ-4
pattern becomes mixed, while AWQ-4 remains predominantly positive across
calibration regimes. This confirms that calibration composition can change
both the magnitude and, in some cases, the direction of the measured
cross-language effect.

\begin{table*}[t]
\centering
\small
\begin{tabular}{llrrrr}
\toprule
Setting & Calibration & Neg. & Similar & Pos. & Range \\
\midrule
GPTQ-4 & Balanced      & 3 & 0 & 1 & $[-2.9,+1.9]$ \\
GPTQ-4 & English only  & 2 & 0 & 2 & $[-3.9,+1.0]$ \\
GPTQ-4 & German only   & 4 & 0 & 0 & $[-6.8,-0.5]$ \\
GPTQ-4 & Bavarian only & 1 & 1 & 2 & $[-1.5,+1.0]$ \\
AWQ-4  & Balanced      & 0 & 1 & 3 & $[0.0,+6.8]$ \\
AWQ-4  & English only  & 0 & 1 & 3 & $[0.0,+6.3]$ \\
AWQ-4  & German only   & 0 & 1 & 3 & $[0.0,+8.7]$ \\
AWQ-4  & Bavarian only & 1 & 0 & 3 & $[-3.4,+5.8]$ \\
\bottomrule
\end{tabular}
\caption{Direction of the per-model Bavarian--German quantisation effect in
the four-model calibration ablation. Negative values indicate greater
Bavarian degradation and positive values greater German degradation.
Values within $\pm0.25$ percentage points are grouped as similar; the final
column reports the per-model BAR--DE range.}
\label{tab:calibration-signs}
\end{table*}

\begin{table*}[t]
\centering
\small
\setlength{\tabcolsep}{5pt}
\begin{tabular}{llrrrr}
\toprule
Model & Quantiser & Balanced & EN-only & DE-only & BAR-only \\
\midrule
EuroLLM 9B    & GPTQ-4 & -2.91 & -2.43 & -6.80 & +0.97 \\
GENBA-10B     & GPTQ-4 & +1.94 & +0.97 & -0.97 & +0.97 \\
Llama 3.1 8B  & GPTQ-4 & -0.49 & +0.97 & -0.49 &  0.00 \\
Qwen2.5 7B    & GPTQ-4 & -1.94 & -3.88 & -4.37 & -1.46 \\
\midrule
EuroLLM 9B    & AWQ-4  & +6.80 & +5.34 & +8.74 & +5.83 \\
GENBA-10B     & AWQ-4  & +6.31 & +6.31 & +2.91 & +3.40 \\
Llama 3.1 8B  & AWQ-4  &  0.00 &  0.00 &  0.00 & -3.40 \\
Qwen2.5 7B    & AWQ-4  & +3.88 & +3.88 & +1.46 & +1.46 \\
\bottomrule
\end{tabular}
\caption{Per-model Bavarian--German dialect-gap amplification
(in percentage points) under different calibration-language
conditions at seed 42. Values are computed as
$\Delta_{\mathrm{BAR}}-\Delta_{\mathrm{DE}}$ relative to
each model's BF16 baseline. Negative values indicate greater
Bavarian degradation, while positive values indicate greater
German degradation. Results are exploratory because calibration
corpora were independently sampled and are not content-aligned.}
\label{tab:calibration-per-model}
\end{table*}

\subsection{Calibration-Source Robustness}
\label{sec:calibration-source}

The main experiments use translated instruction-style calibration data from
Alpaca, Dolly, and OpenHermes. To test whether the observed effects depend on
this type of calibration text, we repeat the 4-bit experiments with natural
Wikipedia text in English, German, and Bavarian. We use the corresponding
November 2023 Wikimedia snapshots and keep the calibration budget and sequence
length unchanged, packing each calibration set into 120 sequences of 2,048
tokens. The corpus is version-pinned and contains no detected exact or near
overlap with \bavground{} or the instruction-style calibration data.

We compare the two calibration sources in two settings. The focused experiment
covers four models, both 4-bit quantisers, all four calibration-language
conditions, and three seeds. We additionally compare balanced calibration
across all 14 models at seed 42. Table~\ref{tab:calibration-source} reports
both comparisons.

\begin{table*}[t]
\centering
\small
\setlength{\tabcolsep}{5pt}
\begin{tabular}{lrrr}
\toprule
Setting & Instruction & Wikimedia & $\Delta$ \\
\midrule
\multicolumn{4}{l}{\emph{Focused: 4 models, 3 seeds, all calibration conditions}} \\
GPTQ-4 & 62.25 & 62.43 & +0.18 \\
AWQ-4  & 61.86 & 63.11 & +1.25 \\
Overall & 62.06 & 62.77 & +0.72 \\
\midrule
\multicolumn{4}{l}{\emph{Finite paired runs: balanced calibration, seed 42}} \\
GPTQ-4 (14 models) & 58.55 & 58.16 & -0.39 \\
AWQ-4 (13 models) & 58.61 & 55.49 & -3.12 \\
Overall (27 pairs) & 58.58 & 56.87 & -1.71 \\
\bottomrule
\end{tabular}
\caption{
\bavground{} accuracy under translated instruction-style and natural
Wikimedia calibration. The focused comparison covers four models, four
calibration-language conditions, two quantisers, and three seeds.
The broader comparison uses balanced calibration at seed 42; the
invalid Gemma~2 AWQ-4 Wikimedia run is excluded from both paired
AWQ conditions. $\Delta$ denotes Wikimedia minus instruction-style
accuracy in percentage points.
}
\label{tab:calibration-source}
\end{table*}

As Table~\ref{tab:calibration-source} shows, changing the calibration source
has little effect on average GPTQ-4 accuracy in the focused three-seed
comparison, with a difference of only $+0.18$ points. AWQ-4 also does not
degrade on average, increasing by $+1.25$ points, but its cross-language
behaviour changes substantially. With Wikimedia calibration, the
Bavarian--German contrast shifts by $-4.63$ points under AWQ-4, as German
improves by $3.76$ points while Bavarian decreases by $0.87$ points. No
corresponding shift is observed for GPTQ-4, where the Bavarian--German
contrast changes by only $+0.56$ points.

The broader comparison shows a different form of source sensitivity.
GPTQ-4 remains close across calibration sources, changing by only
$-0.39$ points across 14 models. Excluding Gemma~2 from both paired
AWQ-4 conditions, AWQ-4 decreases by $3.12$ points across 13 models.
The excluded Wikimedia run produced non-finite option scores for all
618 instances, which were deterministically mapped to option A;
its apparent accuracy is therefore invalid. Aya Expanse and OLMo~2
show large accuracy losses despite finite likelihoods and complete
evaluation runs. Thus, AWQ-4 exhibits greater calibration-source
sensitivity than GPTQ-4 even after excluding the invalid run.

Together, the two comparisons show that calibration source can affect more
than aggregate accuracy. In the replicated focused experiment, changing the
source alters the relative preservation of Bavarian and German under AWQ-4
even though overall accuracy does not decline. In the broader model panel,
the same quantiser also shows stronger model-specific instability. GPTQ-4 is
considerably more stable in both respects. Natural Wikimedia calibration
therefore does not provide a consistent advantage for Bavarian, and the
effect of calibration source depends strongly on the quantisation method.

\subsection{Instability by BF16 Decision Margin}
\label{sec:margin-analysis}
The margin-extreme results are reported in the main text in
Table~\ref{tab:margin-main}. The analysis uses the same paired
BF16--quantised item rows as the transition decomposition and groups examples
by deciles of the BF16 selected-answer margin.

The resulting pattern is strongly monotonic: answer changes are concentrated
among low-margin predictions and become rare for high-margin examples. In the
lowest decile, GPTQ-4 and AWQ-4 alter roughly $40$--$43\%$ of decisions,
whereas only $1$--$2\%$ change in the highest decile. GPTQ-8 is substantially
more stable overall, but still changes $10.16\%$ of the lowest-margin
predictions and none of the highest-margin predictions. This supports the
interpretation that quantisation primarily perturbs decisions that are already
weakly separated in BF16 rather than uniformly disrupting high-confidence
predictions.

\subsection{Item-Level Instability Model}
\label{sec:instability-model}

Table~\ref{tab:instability-model} reports the item-level logistic model
predicting whether quantisation changes the selected answer. The model
includes model and source-question effects; continuous predictors are
standardised, with Bavarian and AWQ-4 as the categorical reference levels.

\begin{table*}[t]
\centering
\small
\begin{tabular}{lrrr}
\toprule
Predictor & Log-odds & OR & $z$ \\
\midrule
BF16 selected-answer margin & -3.263 & 0.038 & -28.18 \\
BF16 correct-answer margin  & -0.040 & 0.961 & -0.92 \\
Prompt tokens               & +0.243 & 1.275 & +1.87 \\
Gold-answer tokens          & -0.549 & 0.577 & -3.80 \\
German vs.\ Bavarian        & -0.065 & 0.938 & -1.13 \\
English vs.\ Bavarian       & +0.002 & 1.002 & +0.02 \\
GPTQ-4 vs.\ AWQ-4           & -0.408 & 0.665 & -8.41 \\
GPTQ-8 vs.\ AWQ-4           & -3.033 & 0.048 & -28.45 \\
\bottomrule
\end{tabular}
\caption{
Item-level model predicting whether quantisation changes the selected answer.
Continuous predictors are standardised. Bavarian and AWQ-4 are the
categorical reference levels. The model includes model and source-question
effects. OR denotes odds ratio.
}
\label{tab:instability-model}
\end{table*}

As shown in Table~\ref{tab:instability-model}, BF16 selected-answer margin is
by far the strongest predictor of instability: a one-standard-deviation
increase corresponds to substantially lower odds of an answer change
($\mathrm{OR}=0.038$). In contrast, the German- and English-versus-Bavarian
coefficients are close to one, providing little evidence that language
identity explains additional item-level fragility once other factors are
accounted for.

\subsection{Item-Level Prediction Instability by Domain}
\label{sec:domain-instability}

\begin{table*}[t]
\centering
\small
\begin{tabular}{llrrrr}
\toprule
Domain & Setting & $\Delta$Acc. & Loss & Gain & Change \\
\midrule
Building \& Sacred Heritage & GPTQ-8 & +0.0 & 0.3 & 0.3 & 1.0 \\
Building \& Sacred Heritage & GPTQ-4 & -2.4 & 5.7 & 3.3 & 12.6 \\
Building \& Sacred Heritage & AWQ-4  & -3.1 & 7.5 & 4.4 & 18.0 \\

Language & GPTQ-8 & -0.4 & 0.5 & 0.1 & 1.0 \\
Language & GPTQ-4 & -2.2 & 5.6 & 3.4 & 15.2 \\
Language & AWQ-4  & -1.9 & 7.3 & 5.4 & 20.7 \\

Politics & GPTQ-8 & -0.3 & 0.8 & 0.5 & 1.5 \\
Politics & GPTQ-4 & +0.4 & 4.4 & 4.8 & 11.9 \\
Politics & AWQ-4  & -0.1 & 5.8 & 5.7 & 16.1 \\

Historical & GPTQ-8 & +0.0 & 0.4 & 0.4 & 1.0 \\
Historical & GPTQ-4 & -2.2 & 5.6 & 3.4 & 11.6 \\
Historical & AWQ-4  & -3.0 & 7.1 & 4.2 & 15.3 \\

Landscape & GPTQ-8 & +0.3 & 0.3 & 0.6 & 1.4 \\
Landscape & GPTQ-4 & -2.4 & 5.1 & 2.8 & 11.0 \\
Landscape & AWQ-4  & -1.6 & 6.6 & 5.0 & 15.0 \\

Living Traditions \& Customs & GPTQ-8 & +0.2 & 0.2 & 0.4 & 0.7 \\
Living Traditions \& Customs & GPTQ-4 & -1.2 & 3.7 & 2.5 & 8.6 \\
Living Traditions \& Customs & AWQ-4  & -0.7 & 4.3 & 3.6 & 10.8 \\

Arts \& Identity & GPTQ-8 & +0.2 & 0.3 & 0.5 & 1.1 \\
Arts \& Identity & GPTQ-4 & -1.6 & 5.3 & 3.7 & 12.4 \\
Arts \& Identity & AWQ-4  & -0.9 & 6.6 & 5.7 & 15.1 \\

Culinary & GPTQ-8 & +0.3 & 0.6 & 0.9 & 1.8 \\
Culinary & GPTQ-4 & +0.0 & 4.7 & 4.7 & 12.3 \\
Culinary & AWQ-4  & +0.1 & 5.5 & 5.6 & 17.0 \\
\bottomrule
\end{tabular}
\caption{
Item-level prediction changes relative to BF16, aggregated by domain.
$\Delta$Acc. is the mean accuracy change in percentage points.
Loss and Gain are correct-to-incorrect and incorrect-to-correct transition
rates (\%), respectively; Change is the overall answer-change rate (\%),
including correctness-preserving answer changes.
}
\label{tab:domain-instability}
\end{table*}

As seen in Table~\ref{tab:domain-instability}, the domain-level transition analysis further illustrates why aggregate
accuracy can understate behavioural change. For example, AWQ-4 changes
$17.0\%$ of Culinary predictions while aggregate Culinary accuracy changes
by only $+0.1$ points. GPTQ-4 similarly changes $11.9\%$ of Politics
predictions despite a small $+0.4$-point accuracy improvement. Large numbers
of losses and gains can therefore cancel at the aggregate level.

\subsection{Most Quantisation-Sensitive Items}
\label{sec:sensitive-items}

\begin{table*}[t]
\centering
\small
\resizebox{\textwidth}{!}{
\begin{tabular}{lllrrrr}
\toprule
Item & Lang. & Domain & Losses & Gains & Loss Rate & Change Rate \\
\midrule
politics\_012 & BAR & Politics & 9 & 2 & 21.4 & 26.2 \\
building\_and\_sacred\_heritage\_025 & BAR & Building/Sacred Heritage & 9 & 0 & 21.4 & 23.8 \\
culinary\_002 & EN & Culinary & 8 & 3 & 19.0 & 28.6 \\
culinary\_022 & EN & Culinary & 8 & 0 & 19.0 & 21.4 \\
language\_024 & DE & Language & 8 & 1 & 19.0 & 21.4 \\
historical\_007 & BAR & Historical & 7 & 5 & 16.7 & 35.7 \\
culinary\_017 & EN & Culinary & 7 & 6 & 16.7 & 33.3 \\
landscape\_025 & DE & Landscape & 7 & 4 & 16.7 & 26.2 \\
politics\_003 & DE & Politics & 7 & 3 & 16.7 & 26.2 \\
language\_006 & DE & Language & 7 & 1 & 16.7 & 23.8 \\
\bottomrule
\end{tabular}
}
\caption{
Ten of the most quantisation-sensitive source-item/language slices, ranked by
loss rate and then answer-change rate across paired quantised comparisons.
Losses count BF16-correct $\rightarrow$ quantised-incorrect transitions;
gains count the reverse. Rates are percentages across all paired
model--quantisation comparisons for that item-language slice.
}
\label{tab:sensitive-items}
\end{table*}

It can be observed in Table~\ref{tab:sensitive-items} that the highest-loss items are not confined to Bavarian. The two largest observed
loss rates occur on Bavarian instances ($21.4\%$), but English and German
items also appear among the most sensitive examples, with loss rates reaching
$19.0\%$. This item-level heterogeneity is consistent with the aggregate
finding that compression does not induce a uniform Bavarian-specific failure
mode.

\subsection{Free-Form Pairwise Judge Results}
\label{sec:freeform-appendix}

\paragraph{Judge protocol.}
The full free-form analysis uses \texttt{gpt-5.6-luna} as a blinded pairwise
judge. Each request provides the language, \bavground{} domain, question, gold
answer, and anonymous Answer A and Answer B. Model names, BF16/quantised status,
and quantiser labels are never shown to the judge. The exact instructions are:
\begin{quote}
\small
You are comparing anonymous short free-form answers to BAVGROUND cultural QA
items. Judge each comparison independently. Prefer the answer that is more
semantically correct with respect to the gold answer. Use ties when both
answers are equivalently correct, equivalently wrong, or indistinguishable in
quality. Do not reward verbosity. Penalize empty/refusal answers and answers
that contradict the gold answer. For Bavarian items, consider dialect
appropriateness, but factual correctness is the primary criterion.
\end{quote}
The structured response contains an overall winner together with separate
fields for factual correctness, instruction following, fluency, and dialect
appropriateness. The paper uses the overall winner as the primary outcome.
Primary comparisons use one randomised side assignment; 20\% are duplicated
with A/B reversed for a position audit, and those audit duplicates are excluded
from ranking aggregates.

The evaluation plan contains 9,270 primary comparisons and
11,115 rows including reversed-order audits. Of these,
9,144 primary comparisons (98.6\%) were successfully parsed;
112 were lost to seven max-output-token truncations and
14 to malformed pair identifiers. Additionally, 1,813
reversed-order judgments were parsed; 1,811 had parsed
primary partners, of which 1,809 were evaluable for the
position-consistency audit after excluding two unjudgeable
pairs. Position consistency was 93.1\% (1,684/1,809).
Missing comparisons are excluded rather than imputed.

\paragraph{Language breakdown.}
Table~\ref{tab:freeform-language} shows that GPTQ-8 is close to BF16 in all
three languages. Four-bit results are more heterogeneous: GPTQ-4 is
comparatively favourable in English but more negative in Bavarian and German,
while AWQ-4 is negative in all three languages. These rates concern relative
preservation against each model's own BF16 answer rather than absolute model
quality.

\begin{table*}[t]
\centering
\small
\begin{tabular}{llrrr}
\toprule
Language & Setting & $N$ & Quant. not worse & Net quant. \\
\midrule
BAR & GPTQ-8 &  997 & 86.1 & -2.5 \\
BAR & GPTQ-4 &  999 & 59.8 & -7.8 \\
BAR & AWQ-4  &  998 & 51.2 & -18.8 \\
DE  & GPTQ-8 & 1021 & 86.8 & -1.3 \\
DE  & GPTQ-4 & 1022 & 59.7 & -13.7 \\
DE  & AWQ-4  & 1021 & 52.2 & -21.7 \\
EN  & GPTQ-8 & 1029 & 87.9 & -3.0 \\
EN  & GPTQ-4 & 1028 & 71.6 & +2.3 \\
EN  & AWQ-4  & 1029 & 59.1 & -12.3 \\
\bottomrule
\end{tabular}
\caption{Full-benchmark free-form preservation by language and
quantiser, using primary comparisons only. Reversed-order
position-audit judgments are excluded. ``Quant. not worse''
is the quantised-win plus tie rate; net quantised preference
is the quantised-win rate minus the BF16-win rate.
Values except $N$ are percentages. The 9,144 primary
comparisons agree with Tables~\ref{tab:freeform-main}
and~\ref{tab:freeform-model}.}
\label{tab:freeform-language}
\end{table*}

\paragraph{Model-level retention.}
Pooling the three quantisers for each model, EuroLLM and Mistral show the
smallest net loss relative to their BF16 outputs, whereas Qwen2.5 and \genba{}
show larger net differences (Table~\ref{tab:freeform-model}). This ranking is
not an absolute model-quality leaderboard.

\begin{table*}[t]
\centering
\small
\begin{tabular}{lrrr}
\toprule
Model & $N$ & Quant. not worse & Net quant. \\
\midrule
EuroLLM 9B & 1850 & 77.2 & -5.8 \\
Mistral 7B & 1824 & 75.3 & -5.9 \\
Llama 3.1 8B & 1798 & 64.3 & -6.9 \\
Qwen2.5 7B & 1835 & 58.1 & -12.1 \\
GENBA-10B & 1837 & 66.3 & -13.0 \\
\bottomrule
\end{tabular}
\caption{Model-level free-form quantisation retention, pooled over GPTQ-8,
GPTQ-4, and AWQ-4. Values except $N$ are percentages.}
\label{tab:freeform-model}
\end{table*}

Across models, GPTQ-8 is the best-retained setting for EuroLLM
($94.2\%$ not worse; $+1.3$ net), Mistral ($91.8\%$; $-1.2$), \genba{}
($88.1\%$; $-1.5$), and Llama~3.1 ($83.6\%$; $-3.0$). Qwen2.5 differs:
its best free-form setting is GPTQ-4 ($54.7\%$ not worse; $-3.6$ net), even
though GPTQ-8 is the most conservative setting for most other models.

\paragraph{Bavarian retention.}
The highest-retained Bavarian model--quantiser combinations are dominated by
GPTQ-8: Mistral GPTQ-8 has a $-0.5$-point net preference, EuroLLM and
\genba{} GPTQ-8 each have $-1.5$ points, and Llama~3.1 GPTQ-8 has $-2.1$
points. EuroLLM GPTQ-4 is the strongest displayed 4-bit Bavarian setting at
$-2.9$ points. This pattern reinforces the main result that 8-bit quantisation
is substantially more stable, while also showing that the remaining
free-form degradation is not unique to Bavarian.

\subsection{Memory and Throughput Measurements}
\label{sec:efficiency-appendix}

Table~\ref{tab:efficiency} reports the representative configurations used for
the quality--memory analysis in Section~\ref{sec:results-efficiency}. The
memory reductions are substantial across both 8-bit and 4-bit settings, but
the batch-size-one throughput measurements do not show a corresponding speed
improvement. In particular, the lower-precision Mistral and EuroLLM
configurations use considerably less memory than their BF16 counterparts while
processing fewer items per second on the inference stack used here.

\begin{table*}[t]
\centering
\small
\setlength{\tabcolsep}{4pt}
\begin{tabular}{llrrrr}
\toprule
Model & Setting & Acc. & Mem. & Red. & Items/s \\
\midrule
EuroLLM 9B        & BF16   & 68.9 & 23.2 &  0 & 2.42 \\
EuroLLM 9B        & GPTQ-8 & 69.4 & 14.2 & 38 & 0.98 \\
EuroLLM 9B        & GPTQ-4 & 65.2 & 10.2 & 56 & 1.31 \\
\midrule
Mistral 7B v0.3   & BF16   & 66.8 & 19.2 &  0 & 3.11 \\
Mistral 7B v0.3   & GPTQ-8 & 66.3 & 11.4 & 41 & 1.69 \\
Mistral 7B v0.3   & GPTQ-4 & 67.2 &  7.9 & 59 & 1.70 \\
\midrule
Qwen2.5 7B        & GPTQ-4 & 65.4 &  9.1 & 54 & 2.06 \\
\bottomrule
\end{tabular}
\caption{Representative quality, memory, and throughput measurements.
Acc. is \bavground{} option-text accuracy (\%); Mem. is peak reserved GPU
memory in GB; Red. is the reduction relative to the BF16 version of the same
model; and Items/s is batch-size-one evaluation throughput on the
GH200/Transformers inference setup.}
\label{tab:efficiency}
\end{table*}

These measurements show that the main deployment benefit in this setting is
capacity rather than inference speed. The result is specific to the hardware,
software stack, and batch-size-one regime used here, and should not be taken
to imply that quantisation cannot improve throughput under other inference
engines or deployment conditions.

Across all profiled configurations, 11 quantised settings reduce peak reserved
memory by at least $50\%$ while remaining within one percentage point of their
BF16 accuracy; 15 remain within two points at the same memory-reduction level.
The largest reduction within the one-point range is Mistral~7B GPTQ-4 at
$58.9\%$. Comparing GPTQ precisions within models also shows that quality is
not monotonic in bit-width: GPTQ-4 scores below GPTQ-8 for 9 of 14 models,
but matches or exceeds it for five. Table~\ref{tab:efficiency-summary}
summarises these quality--memory trade-offs across the profiled configurations.
These comparisons are descriptive and should not be interpreted as evidence
that 4-bit quantisation generally improves quality.

\begin{table*}[t]
\centering
\small
\begin{tabular}{lr}
\toprule
Criterion & Configurations \\
\midrule
$\geq 50\%$ memory reduction, within 1 pp of BF16 & 11 \\
$\geq 50\%$ memory reduction, within 2 pp of BF16 & 15 \\
Quantised accuracy $\geq$ BF16 accuracy & 19 \\
GPTQ-4 accuracy $<$ GPTQ-8 & 9/14 \\
GPTQ-4 accuracy $\geq$ GPTQ-8 & 5/14 \\
\bottomrule
\end{tabular}
\caption{Summary of quality--memory trade-offs across the profiled
configurations. Accuracy comparisons are descriptive; ``within 1 pp'' and
``within 2 pp'' refer to absolute percentage-point differences from each
model's BF16 accuracy.}
\label{tab:efficiency-summary}
\end{table*}


\section{Experimental setup}
\label{tab:experimental-setup-appendix}

\subsection{Hardware and Software}
\label{tab:experimental-setup-appendix-hw}

The metadata captured by the completed runs reports the environment summarised
in Table~\ref{tab:appendix-env}. We omit secrets and user-specific cache paths
from the paper, but preserve the relevant hardware and package versions needed
to interpret the measurements.

\begin{table*}[htbp]
\centering
\small
\begin{tabular}{p{0.28\linewidth}p{0.62\linewidth}}
\toprule
\textbf{Component} & \textbf{Captured setting} \\
\midrule
Accelerator & NVIDIA GH200 reported by \texttt{nvidia-smi}; 97,871 MiB device memory (approximately 95.6 GiB), driver 580.95.05, and 900 W power limit. \\
Operating system & Linux on \texttt{aarch64}; kernel \texttt{6.2.0-1015-nvidia-64k}. \\
Python & Python 3.10.12. \\
Core packages & \texttt{torch} 2.13.0, \texttt{transformers} 5.14.1, \texttt{accelerate} 1.14.0, \texttt{GPTQModel} 7.3.2, \texttt{bitsandbytes} 0.50.0, \texttt{numpy} 2.2.6, \texttt{pandas} 2.3.3, \texttt{scipy} 1.15.3, \texttt{nvidia-ml-py} 13.610.43. \\
Inference backend & Hugging Face Transformers with automatic device mapping. Quantised checkpoints are loaded as Hugging Face-compatible GPTQ or AWQ checkpoints. \\
\bottomrule
\end{tabular}
\caption{Hardware and software details captured in run metadata.}
\label{tab:appendix-env}
\end{table*}

\subsection{Evaluation, Calibration, and Runtime Configuration}
Table~\ref{tab:appendix-runtime} summarises the common runtime configuration.
The full evaluation uses 618 \bavground{} instances and three protocols, so a
completed model--precision run produces 1,854 result rows. The reported matrix
contains 56 completed model--precision runs, corresponding to 103,824 raw
result rows before paper-specific aggregation.

\begin{table*}[htbp]
\centering
\small
\begin{tabular}{p{0.28\linewidth}p{0.62\linewidth}}
\toprule
\textbf{Aspect} & \textbf{Setting} \\
\midrule
Evaluation config & \texttt{configs/evaluation/bavground\_full.yaml}; languages \texttt{en}, \texttt{de}, and \texttt{bar}; protocols \texttt{option\_text\_avg}, \texttt{letter}, and \texttt{generate\_letter\_and\_text}. \\
Runtime parameters & Batch size 1, maximum sequence length 4096, maximum generation length 24 tokens, random seed 42, prompt saving enabled, and resumable runs enabled. \\
Quantisation settings & BF16 reference plus GPTQ-8, GPTQ-4, and AWQ-4. Quantised runs use group size 128, asymmetric zero-point quantisation where supported, and source precision for embeddings and the language-model head. \\
Calibration & Balanced English--German--Bavarian calibration condition, seed 42, target size approximately 250k tokens, and calibration sequence length 2048. No exact normalised-text overlap was detected with \bavground{} questions, answer options, or recorded source references. \\
Runtime telemetry & Run metadata records model-loading duration, peak allocated and reserved GPU memory after loading and inference, average and maximum power draw, total energy, phase-level energy for model loading and inference, and \texttt{nvidia-smi} snapshots before and after execution. \\
\bottomrule
\end{tabular}
\caption{Common evaluation, calibration, and runtime settings.}
\label{tab:appendix-runtime}
\end{table*}


\section{Reproducibility Details}
\label{sec:reproducibility}

\subsection{Model Checkpoint Identifiers}
\label{app:model-checkpoints}

Table~\ref{tab:checkpoint-identifiers} lists the checkpoint identifiers used
for the models in Table~\ref{tab:main-results}. The \genba{} instruction checkpoint is
reported under an anonymised local identifier during double-blind review; the
released artifact records its immutable revision/hash without exposing a
user-specific filesystem path.

\subsection{Quantisation Parameters}
\label{sec:quant-parameters}

Table~\ref{tab:quant-parameters} summarises the common quantisation settings
used throughout the main 14-model matrix. GPTQ-8 and GPTQ-4 differ only in
weight precision, while GPTQ-4 and AWQ-4 provide the corresponding
method-level comparison at 4-bit precision. Unless otherwise noted, all
quantised checkpoints use the same group size, calibration budget, sequence
length, and balanced English--German--Bavarian calibration condition.

\begin{table*}[t]
\centering
\small
\begin{tabular}{ll}
\toprule
Parameter & Main-matrix setting \\
\midrule
GPTQ weight precision & 8-bit or 4-bit \\
AWQ weight precision & 4-bit \\
Group size & 128 \\
Calibration budget & $\sim$250k tokens \\
Calibration sequence length & 2048 \\
Calibration seed & 42 \\
Calibration languages & Balanced EN/DE/BAR \\
Embeddings & Source dtype \\
LM head & Source dtype \\
GPTQ inference dtype & FP16 \\
\bottomrule
\end{tabular}
\caption{
Common quantisation settings used in the main evaluation matrix.
Backend-specific saved checkpoint settings and any architecture-dependent
exceptions are reported in the released run manifest.
}
\label{tab:quant-parameters}
\end{table*}

As shown in Table~\ref{tab:quant-parameters}, the main matrix holds
calibration composition and core quantisation settings fixed across models so
that observed differences primarily reflect model, bit-width, and quantiser
effects rather than deliberately changing the calibration regime. The
calibration-language ablation in Section~\ref{sec:calibration-ablation}
relaxes only the language composition of the calibration corpus while keeping
the remaining settings fixed.

\end{document}